%% file: conference_101719.tex
\documentclass[conference]{IEEEtran}
\IEEEoverridecommandlockouts
\usepackage{cite}
\usepackage[T1]{fontenc}
\usepackage{enumitem}
\usepackage{amsmath,amssymb,amsfonts}
\usepackage{algorithmic}
\usepackage{graphicx}
\usepackage{textcomp}
\usepackage{xcolor}
\usepackage{booktabs}
\usepackage{hyperref}

\def\BibTeX{{\rm B\kern-.05em{\sc i\kern-.025em b}\kern-.08em
    T\kern-.1667em\lower.7ex\hbox{E}\kern-.125emX}}
\begin{document}

\title{Fed-ReMasker: Federated Tabular Imputation under Feature-Level Missingness}

\author{\IEEEauthorblockN{Ioannis Papathanail\IEEEauthorrefmark{1}\IEEEauthorrefmark{2}, 
Rooholla Poursoleymani\IEEEauthorrefmark{1}\IEEEauthorrefmark{2}\IEEEauthorrefmark{3}, 
Lubnaa Abdur Rahman\IEEEauthorrefmark{2}\IEEEauthorrefmark{3}, and 
Stavroula Georgia Mougiakakou\IEEEauthorrefmark{2}, \\
\textit{on behalf of the BETTER4U project consortium}}

\IEEEauthorblockA{\IEEEauthorrefmark{1}These authors contributed equally to this work.}

\IEEEauthorblockA{\IEEEauthorrefmark{2}\textit{ARTORG Center for Biomedical Engineering Research}\\
\textit{University of Bern}, Bern, Switzerland}

\IEEEauthorblockA{\IEEEauthorrefmark{3}\textit{Graduate School for Cellular and Biomedical Sciences}\\
\textit{University of Bern}, Bern, Switzerland}
}

\maketitle

\begin{abstract}
Multi-center clinical studies and biomedical research collaborations increasingly seek to utilize data across centers to build models that generalize beyond any single center. 
This creates two distinct challenges: data protection regulations may restrict the sharing of raw patient data across institutions, while centers may collect only partially overlapping sets of features under different protocols.
Federated learning enables collaborative model training without centralizing raw data. 
However, existing federated imputation methods rarely evaluate feature-level missingness, in which entire features are unobserved at some centers.
To address this setting, we adapt the ReMasker masked autoencoder to federated learning (Fed-ReMasker), enabling centers to impute features never observed locally by leveraging knowledge learned across collaborating centers. 
We evaluate Fed-ReMasker in a benchmark spanning synthetic datasets with linear and nonlinear relationships and real-world tabular datasets, including clinical data. 
The benchmark varies the number of centers, the missingness ratios, and client heterogeneity. 
Fed-ReMasker achieves the lowest imputation error in 93.2\% of value-level and 96.7\% of feature-level scenarios in the homogeneous benchmark.
It also remains robust to client heterogeneity using simple federated averaging, outperforming all baselines in all 36 value-level scenarios and each baseline in at least 35 of 36 feature-level scenarios, and comes within 3.0\% on average of a centralized model trained on the pooled data.
\end{abstract}


\section{Introduction}
 
Tabular data are among the most common modalities in machine learning and are widely used across healthcare, biomedical research, and other data-driven applications \cite{jiang2026representation}. 
Increasingly, they are collected independently by centers that seek to collaborate but cannot share raw records due to data protection regulations, legal restrictions, or institutional governance policies \cite{kairouz2021advances,zhang2024recent}. 
Federated Learning (FL) addresses this by enabling centers to train shared models while keeping their records local, making it a widely used approach for privacy-preserving collaborative training \cite{mcmahan2017communication,rieke2020future}.
 
A major challenge in real-world tabular data is the presence of missing values.
Missingness can occur for many reasons, such as incomplete measurements, optional questionnaires, device failures, differences in data collection protocols, or limited resources at some sites \cite{haneuse2021assessing, nguyen2024fedmac}.
Many imputation methods have been developed for centralized settings, ranging from classical statistical approaches to deep generative and masked-autoencoding models \cite{van2011mice,stekhoven2012missforest,yoon2018gain,mattei2019miwae,du2024remasker}.
In federated settings, missing-data imputation is receiving increasing attention, with recent works exploring generative models, personalized aggregation strategies, and adversarial learning without requiring raw-data centralization \cite{balelli2023fed,min2025cafe,li2025fedimpute,hocine2026federated}.

In multi-center clinical studies, centers often collect data under different protocols and measure different feature sets, so a feature recorded at some centers may be entirely absent at others \cite{resche2013multiple,hughes2021combining}. 
This results in feature-level missingness, where a client has no local observations of a given feature.
Unlike value-level missingness, where some values of the feature remain available locally, this setting provides no local information from which to learn its distribution or relationships with other variables.
In this work, we investigate federated tabular imputation under feature-level missingness, enabling multi-center studies to leverage such incompatible datasets collaboratively.
 
We consider a horizontal federated setting with a shared global feature schema: each center is treated as a client, and some features may be unobserved at some clients.
Missingness mechanisms are commonly classified as Missing Completely At Random (MCAR), Missing At Random (MAR), or Missing Not At Random (MNAR), according to whether missingness is independent of the data, depends only on observed values, or depends on unobserved values \cite{little2019statistical}.
To isolate the effect of feature-level missingness from more complex missing mechanisms, we focus on MCAR missingness at both the value and feature level.
We additionally assess robustness beyond MCAR with a heterogeneity condition in which clients are formed by clustering in feature space, so the feature-level missingness becomes data-dependent (Section~\ref{sec:hetero}).

Our contributions are as follows:

\begin{itemize}

\item We adapt the ReMasker masked autoencoder \cite{du2024remasker} to this federated setting (Fed-ReMasker), enabling clients to impute features never observed locally, and show that it achieves the lowest error among existing federated imputation methods, with the largest gains on entirely missing features.

\item  To the best of our knowledge, we provide the first systematic benchmark of federated tabular imputation under both value- and feature-level missingness, explicitly including features entirely unobserved at individual clients, and evaluated across client counts, synthetic and real datasets, and four client-heterogeneity conditions.

\item For Fed-ReMasker, we compare federated optimization strategies (FedAvg, FedProx, FedAdam) and quantify the cost of federation against a centralized reference.
\end{itemize}

Our experiments show that Fed-ReMasker outperforms existing FL imputation methods across datasets under both value- and feature-level missingness. 
It also remains robust under client heterogeneity, with standard FedAvg aggregation performing competitively with more elaborate strategies.

\section{Related Work}
\label{sec:relatedwork}
 
\subsection{Centralized Imputation}
Missing-value imputation has been widely studied in centralized settings, from classical methods (mean/median imputation, K-Nearest Neighbors, Multivariate Imputation by Chained Equations (MICE) \cite{van2011mice}, and MissForest \cite{stekhoven2012missforest}) to deep learning approaches.
The Missing data Importance-Weighted Autoencoder (MIWAE) \cite{mattei2019miwae} uses a deep latent variable model and optimizes an importance-weighted lower bound on the observed-data likelihood.
Generative Adversarial Imputation Nets \cite{yoon2018gain} formulate imputation as a generative adversarial learning problem, where a generator imputes missing values and a discriminator distinguishes observed from imputed entries.
ReMasker \cite{du2024remasker} extends masked auto-encoding to tabular imputation: during training, it masks a subset of observed values and learns to reconstruct them, and at inference, it imputes genuinely missing values.
These methods are effective but assume that all data are centrally accessible.
 
\subsection{Federated Imputation}
Several FL methods address incomplete or heterogeneous client feature spaces without directly imputing missing values:
FLIC \cite{rakotomamonjy2023personalised} learns a shared latent representation across heterogeneous feature spaces, DARN \cite{zhang2025federated} uses a missing-aware transformer for incomplete-data prediction, and FedFeatGen \cite{poudel2025multimodal} reconstructs missing modalities in representation space.
These methods focus on representation learning or downstream prediction rather than data-level imputation.
 
Within this broader literature, several methods specifically address federated imputation for tabular data.
Fed-MIWAE \cite{balelli2023fed} extends MIWAE to the federated setting and supports imputation under MCAR and MAR using Federated Averaging (FedAvg).
During inference, missing values are imputed as a weighted average of decoder samples based on how well they explain the observed data.
Complementarity Adjusted FEderated averaging (Cafe) \cite{min2025cafe} focuses on missing-data heterogeneity across clients.
For each feature, it learns an imputation model from the other features and a logistic model of its missingness mechanism.
These models are used to compute pairwise complementarity scores between clients, which are then combined with sample-size weights to produce personalized models for each client.
FedImpute \cite{li2025fedimpute} uses a two-phase approach that separates samples with low and high missingness. 
It then uses a generative adversarial network with clustering and auxiliary models to improve imputation for highly missing samples.
These methods address federated imputation for tabular data, but mainly consider value-level or heterogeneous missingness patterns, rather than settings in which entire features are unobserved at individual clients.

FedHF-Impute \cite{hocine2026federated} addresses federated imputation under heterogeneous feature spaces by constructing a global feature graph from client-level feature correlations.
It trains a graph neural network to impute missing values using information from related observed features.
Its evaluation introduces additional value-level missingness only among entries that are originally available, and computes error exclusively on those artificially corrupted positions, so it does not include positions belonging to features that are unobserved at a client.
It evaluates a fixed setting with four clients, in which 60\% of features are available at each client.
In contrast, we evaluate both value- and feature-level missingness across different client counts and missingness levels, on both synthetic and real datasets.

Closer to our setting, cross-site imputation \cite{thiesmeier2025cross} addresses variables entirely unrecorded at some sites without pooling raw data.
It transports regression coefficients across sites within a multiple-imputation framework. 
Unlike our benchmark, it targets valid inference in a downstream analysis rather than point-accurate imputation, and is formulated for a single unrecorded variable at a time, so it is not directly comparable under Normalized Root Mean Squared Error (NRMSE).

\subsection{Federated Optimization}
Beyond the choice of imputation method, federated optimization can also influence performance, particularly when client data are heterogeneous.
FedAvg \cite{mcmahan2017communication} averages client updates by sample size.
FedProx \cite{li2020federated} adds a proximal term to limit client drift during local training.
Adaptive server methods such as FedAdam \cite{reddi2020adaptive} instead apply a stateful optimizer to the aggregated update. 
We therefore compare these strategies in Section~\ref{sec:aggregation} to assess whether more specialized optimization improves federated imputation under client heterogeneity.

\section{Materials and Methods}
 
\subsection{Problem Formulation}
We consider a horizontal FL problem with a shared global feature schema. 
Clients train locally, and a central server aggregates their updates, so only model parameters are communicated.
The underlying data are represented by $X\in\mathbb{R}^{N\times d}$, where $N$ is the total sample size and $d$ is the number of features in the global schema. 
Individual clients may have entire features unobserved, but all clients use the same global feature indexing.
Each client $k \in \{1, \dots, K\}$ holds a local dataset in $X^{(k)}\in\mathbb{R}^{N_k\times d}$, where $N_k$ is its sample size. 
Each client also has a missingness mask $M^{(k)} \in \{0,1\}^{N_k \times d}$, where $M^{(k)}_{ij} = 0$ indicates that the value of sample $i$ at feature $j$ is not observed.
We define \emph{feature-level missingness} as the case where $M^{(k)}_{ij} = 0$ for every local sample $i$ at feature $j$.
Thus, feature $j$ is entirely unobserved at client $k$. 
In contrast, \emph{value-level missingness} refers to cases where only individual entries of a feature are missing. Importantly, feature- and value-level missingness are applied jointly, such that a feature entirely absent at one client may be only partially observed at other clients.  
The goal is to estimate each missing entry $\hat{x}_{ij}$ as close as possible to its unobserved ground truth $x_{ij}$, for both value- and feature-level missingness.
Value-level missingness is MCAR in all scenarios.
Feature-level missingness is likewise MCAR wherever clients are partitioned at random, since the features removed from a client are drawn independently of the data.
The exception is one of the heterogeneity conditions, i.e., the feature-distribution skew (Section~\ref{sec:hetero}), where clients are formed by clustering in feature space, so the feature-level mask is no longer marginally independent of the feature values.
Conditional on client membership, the feature-level mask is generated independently of the feature values. 
Marginally, however, the mask is dependent on the feature values since client membership is itself determined from the feature space. 
We therefore treat this condition as data-dependent missingness rather than MCAR.

\subsection{Fed-ReMasker}
Fed-ReMasker is our federated adaptation of the masked autoencoder ReMasker \cite{du2024remasker}, in which each client trains a local ReMasker on its observed data.
We keep ReMasker's architecture and masked-reconstruction objective unchanged from the original.
Fed-ReMasker differs from ReMasker only in the training setting, whereby it is trained across distributed clients rather than centrally.
After local training, the client model parameters are aggregated via FedAvg, weighted by local dataset size.
For a feature that is absent at client $k$, no local imputation targets are available.
However, clients that observe the feature can learn from and update the shared model parameters using that feature.
Their updates are incorporated into the global model through federated aggregation. 
The resulting global model can transfer information learned from that feature to clients where it is unobserved.
At inference, each client applies the final global model to impute both missing values and entirely missing features.

\subsection{Federated Optimization Strategies}
\label{sec:aggregation}

Let $w_t$ denote the global model at round $t$, let $w_t^{k}$ denote the parameters returned by client $k$ after local training in round $t$, and let $F_k$ denote its local loss.
FedAvg \cite{mcmahan2017communication} sets the next global model to a sample-size-weighted average:
\begin{equation}
\label{eq:fedavg}
w_{t+1}=\sum_k p_k w_t^{k}
\end{equation}
where $p_k=N_k/N$.
Writing the aggregated client update as $\Delta_t=\sum_k p_k\left(w_t^{k}-w_t\right)$, \eqref{eq:fedavg} is equivalently a unit server step, $w_{t+1}=w_t+\Delta_t$.

We also evaluate FedProx \cite{li2020federated} and FedAdam \cite{reddi2020adaptive} as alternative optimization strategies. 
FedAdam is an Adam-based FedOpt variant designed to improve robustness under client heterogeneity.
FedProx adds a proximal term to each client's local objective,
\begin{equation}
\label{eq:fedprox}
\min_w \; F_k(w) + \tfrac{\mu}{2}\lVert w - w_t\rVert^2 ,
\end{equation}
where the coefficient $\mu\geq0$ controls the strength of the penalty on deviation from the current global model $w_t$.
The proximal term in \eqref{eq:fedprox} limits client drift during local training.
When $\mu=0$, FedProx reduces to FedAvg.

FedAdam leaves local training unchanged and instead applies a stateful Adam-based server optimizer to the aggregated client update $\Delta_t$.
It maintains first- and second-moment estimates $m_t$ and $v_t$, and updates the global model $w_{t+1}$ as:

\begin{equation}
\label{eq:fedadam}
\begin{split}
& m_t = \beta_1 m_{t-1} + (1-\beta_1)\Delta_t \\
& v_t = \beta_2 v_{t-1} + (1-\beta_2)\Delta_t^{2} \\
& w_{t+1} = w_t+\eta\ \frac{m_t}{\sqrt{v_t}+\tau} 
\end{split}
\end{equation}

In \eqref{eq:fedadam}, $\eta$ is the server learning rate, $\beta_1,\beta_2\in[0,1)$ are the moment decay rates, and $\tau$ is a small constant for numerical stability. 
The FedAdam server learning rate $\eta$ and the FedProx proximal coefficient $\mu$ are selected as described in Section~\ref{sec:hetero}.

\section{Experimental Setup}
 
\subsection{Datasets}
 
To benchmark the federated imputation, we use both synthetic and real datasets.
We use three open-access real-world datasets:
\begin{itemize}
\item \textit{Codon} \cite{codon_usage_577}, a codon usage frequency dataset.
We retain only genomic DNA entries, so that each species contributes a single row rather than separate genomic, mitochondrial, and chloroplast entries, and exclude plasmid records.
We drop five identifier and metadata columns and retain only fully observed rows, so ground truth exists at every evaluation position.
This results in 9{,}249 samples and 64 codon-frequency features.

\item \textit{PhysioNet} \cite{goldberger2000physiobank,silva2012predicting}, a clinical intensive care unit time-series dataset, is converted to a patient-level tabular representation by retaining the last measurement of each parameter per patient. 
We remove categorical features, features with more than 20\% missingness, and samples with remaining missing values. 
This results in 9{,}313 samples and 19 numeric features.

\item \textit{NHANES} \cite{nhanes_2003_2018}, a population health survey, is constructed by merging the 2003--2018 cycles at the participant level.
After applying similar filtering (categorical removal, $\leq$30\% missingness, complete-case samples), 36{,}758 samples and 32 numeric features remain.
\end{itemize}

For the synthetic datasets, we use the Causal Discovery Toolbox \cite{kalainathan2019causal} to generate six datasets with known causal structure.
Three datasets have linear relationships, and three use neural-network-induced nonlinear relationships. 
For each type, we generate one dataset for each $d\in\{20,50,100\}$.
Each dataset contains 50{,}000 samples and additive Gaussian noise with coefficient 0.4, following \cite{huang2023towards}.

Global feature-wise statistics (mean, minimum, and range) are aggregated from local client statistics without accessing raw data. 
The global mean is used for mean imputation and to initialize Cafe, and the global minimum and range are used to normalize each feature to the interval $[0, 1]$ using min-max scaling.
Missingness masks are applied before these statistics are computed. 
Thus, all the preprocessing statistics are derived only from observed values.
Ground truth values at masked positions, including positions belonging to entirely missing features, are used solely for evaluation, never during preprocessing or training.
We consider FL without secure aggregation or differential privacy.
Accordingly, the benchmark evaluates decentralized imputation without raw data sharing, rather than providing formal privacy guarantees.

\subsection{Missingness Simulation and Federated Setup}

We control feature-level missingness through a Missing Feature Ratio (MFR) and value-level missingness through a Missing Value Ratio (MVR).
We refer to each combination of dataset, client count, MFR, and MVR as a scenario.
In the heterogeneity study, a scenario additionally specifies the heterogeneity condition and partition seed.

\subsubsection{Homogeneous client scenarios}
We introduce feature-level missingness by removing a fraction of features per client ($\text{MFR} \in\{0, 0.1, 0.2, 0.3\}$), then apply additional random value-level missingness ($\text{MVR} \in\{0.1, 0.3, 0.5\}$) to the remaining observed values.
Each dataset is split into $K \in \{3, 5, 10\}$ clients, with each client assigned at least $N/(2K)$ samples, where $N$ is the dataset size.
For each feature, one client is randomly designated to retain that feature and is not allowed to remove it.
Each client then removes $\left\lceil \mathrm{MFR} \cdot d \right\rceil$ features, sampled uniformly without replacement from the features it is allowed to remove.
Every feature is therefore observed by at least one client, while a feature may be dropped by several.
Since the number of removed features is rounded up, the realized per-client rate is $\left\lceil \mathrm{MFR} \cdot d \right\rceil / d$, which exceeds the nominal MFR when $\mathrm{MFR} \cdot d$ is not an integer.
We refer to these randomly partitioned clients, which share the same MFR and MVR, as the homogeneous setting, although clients still differ in which features they observe.
This yields 216 synthetic (6 datasets $\times$ 4 MFR $\times$ 3 MVR $\times$ 3 $K$) and 108 real-data scenarios (3 datasets $\times$ 4 MFR $\times$ 3 MVR $\times$ 3 $K$).
All homogeneous scenarios use a single fixed partition seed, since the grid already spans 324 configurations.
 
\subsubsection{Heterogeneous client scenarios}
\label{sec:hetero}
Beyond the homogeneous grid, we construct four heterogeneity conditions on the three real datasets.
We fix $\text{MFR}=0.2$, $\text{MVR}=0.3$, and $K=5$, so that the average missingness rates remain constant while how samples and missingness are distributed across clients changes.
As in the homogeneous scenarios, we ensure that every feature is observed by at least one client.
\begin{itemize}
\item \textit{Quantity skew} creates unequal client sample sizes with target Gini coefficients of 0.2 and 0.35.
We use a geometric size profile and randomly assign samples to fill the resulting client sizes. 
\item \textit{Missingness skew} assigns clients 1--5 MFRs of 0.1, 0.15, 0.2, 0.25, and 0.3, respectively, instead of a uniform value.
Thus, clients differ in how many features they observe, while which features each client removes is drawn by the same procedure as in the homogeneous setting.
\item \textit{Feature-distribution skew (non-IID)} partitions samples using $k$-means clustering on the complete pre-masking feature matrix after standardization, which is used only to construct the benchmark partition and is never accessible to the federated algorithms.
The strength of the dependence between client membership and feature values depends on each dataset's correlation structure.
To isolate feature-distribution from quantity skew, we cluster into $2K$ groups and greedily assign clusters to clients to balance sample counts, so client sizes remain comparable despite the non-IID partition.
\end{itemize}

The two Gini levels for quantity skew yield two conditions, giving four heterogeneous conditions in total: quantity skew (Gini 0.2), quantity skew (Gini 0.35), missingness skew, and feature-distribution skew.
Each condition uses three independent partition seeds.
We run the same three-seed protocol on one homogeneous setting ($\text{MFR}=0.2$, $\text{MVR}=0.3$, $K=5$) as a reference.
At a given seed, the quantity-skew and feature-distribution-skew conditions remove the same features from each client index as this homogeneous reference, so differences between them are attributable to the partition alone.
Missingness skew necessarily differs, since it varies how many features each client loses.
This yields 36 heterogeneous scenarios (3 datasets $\times$ 4 conditions $\times$ 3 seeds) and 9 homogeneous reference scenarios (3 datasets $\times$ 3 seeds).

We additionally compare the FedAvg, FedProx, and FedAdam optimization strategies for Fed-ReMasker on the 45-scenario robustness subset described above.
To select $\mu$ for FedProx and the server learning rate $\eta$ for FedAdam, we use leave-one-dataset-out hyperparameter selection over the same grid $\{0.001,0.01,0.1,1.0\}$ for both.
For each held-out real dataset, we select the value that minimizes the mean overall NRMSE across the other two real datasets, and apply it unchanged across all conditions for that dataset.

\subsubsection{Centralized reference}
\label{sec:central_setup}
To quantify the performance cost of federation, we additionally train a centralized ReMasker on the pooled client data, using the same architecture and optimizer settings and an equal number of data passes ($T \cdot E = 300$ epochs).
Pooling converts feature-level missingness into value-level missingness: a feature absent at some clients becomes a column with missing values for those clients' samples, in addition to the value-level missingness applied to the rest.
The centralized model therefore never encounters an entirely unobserved feature, because each column is retained by at least one client.
The overall missingness rate is also the same for every client count, because each client always removes the same number of features.
Quantity skew and per-client MFR spread leave the mask independent of feature values, so pooling reduces them to a value-independent pooled mask, whereas under feature-distribution skew, the clients are feature-space clusters and the correlation between the mask and the feature values survives pooling.
The centralized reference is evaluated on all nine datasets at $\text{MFR}\in\{0,0.1,0.2,0.3\}$ and $\text{MVR}\in\{0.1,0.3,0.5\}$ with $K=5$ (9 $\times$ 4 $\times$ 3 = 108 scenarios).
Because the pooled missingness rate is invariant to $K$, we sweep the client count only at the central operating point $\text{MFR}=0.2$, $\text{MVR}=0.3$, adding $K\in\{3,10\}$ (9 $\times$ 2 = 18), for 126 scenarios in total.
It is also run under the feature-distribution skew condition and the homogeneous reference of the heterogeneity study, with three seeds on the three real datasets.

\subsection{Baselines}
We evaluate Fed-ReMasker against four methods: Mean imputation (a non-learnable baseline using the global mean), Cafe \cite{min2025cafe}, Fed-MIWAE \cite{balelli2023fed}, and FedHF-Impute \cite{hocine2026federated}.
These methods are described in Section~\ref{sec:relatedwork}.
For features entirely missing at a client, Cafe falls back to a sample-size-weighted average of models from clients that observed the feature.
Fed-MIWAE has no local imputation signal for such features but can still impute them at inference using the global model.
FedHF-Impute imputes such features from related observed features through its global feature graph.

\subsection{Evaluation Metrics}
For each scenario, we generate missingness masks from the complete dataset. 
Each client trains only on its resulting observed entries and imputes the masked positions in the same rows. 
During training, methods may generate internal masking or corruption patterns as specified by their respective algorithms.
These training masks are independent of the evaluation mask.
The original values at masked positions are retained solely as ground truth for evaluation and are never used in preprocessing, normalization, or training.

We compute imputation quality using NRMSE. 
For client $k$:
\begin{equation}
\label{eq:nrmse_client}
\text{NRMSE}_k = \sqrt{\frac{1}{|\mathcal{M}_k|} \sum_{(i,j) \in \mathcal{M}_k} \left(\frac{\hat{x}_{ij} - x_{ij}}{r_j}\right)^2}
\end{equation}

In \eqref{eq:nrmse_client}, $\mathcal{M}_k$ denotes the set of missing positions $(i, j)$ for client $k$, and $\hat{x}_{ij}$ and $x_{ij}$ are the imputed and ground truth values, respectively.
The normalizer $r_j$ is the range of feature $j$ over all values observed across clients in a scenario. 
It therefore never uses held-out ground truth and is identical for all methods compared within a scenario.

We report NRMSE separately for value-level missing positions ($\text{NRMSE}_\text{val}$) and feature-level missing positions ($\text{NRMSE}_\text{feat}$), as well as for all missing positions.
For each scenario, we compute NRMSE by pooling the squared normalized errors across all missing positions and clients: 

\begin{equation}
\label{eq:nrmse_scenario}
\text{NRMSE}
=
\sqrt{
\frac{
\sum_k \sum_{(i,j)\in\mathcal{M}_k}
\left(
\frac{\hat{x}_{ij}-x_{ij}}{r_j}
\right)^2
}{
\sum_k |\mathcal{M}_k|
}
}.
\end{equation}
We report the unweighted mean of scenario-level NRMSE for each dataset, with feature-level NRMSE computed only for scenarios with $\text{MFR}>0$.
In the heterogeneity experiments, we additionally report equal-client macro NRMSE to assess sensitivity to client-size weighting.
 
As we evaluate all methods on the same scenarios, we also report paired differences, computed for each baseline as $\text{NRMSE}_\text{baseline} - \text{NRMSE}_\text{Fed-ReMasker}$, so that positive values favor Fed-ReMasker.
We summarize these differences using their mean, a 95\% $t$-interval, and the fraction of scenarios in which Fed-ReMasker achieves lower NRMSE.
Because scenarios drawn from the same dataset are not independent, we compute the $t$-interval across datasets, taking each dataset's mean difference as one observation, so the resulting intervals quantify variability across datasets rather than repeated client partitioning.
We assess repeated partitioning separately in the heterogeneity study, where we generate three independent partitions for each condition.
Feature-level comparisons are computed over the 243 scenarios with $\text{MFR}>0$, since the $\text{MFR}=0$ cells contain no feature-level missing positions.

\subsection{Implementation Details}
 \label{sec:impl}
To ensure a common training budget, all methods are trained for $T=60$ federated rounds with 5 local epochs per round.
Baseline hyperparameters follow the corresponding reference implementations, with Fed-MIWAE using the implementation provided in the Cafe repository \cite{min2025cafe}.
For Fed-ReMasker, the masked autoencoder has an encoder depth of 8, a decoder depth of 4, an embedding dimension of 64, and 4 attention heads, with a masking ratio of 0.3.
Clients train on shuffled mini-batches of size 64 using AdamW with $\beta_1{=}0.9$, $\beta_2{=}0.95$, a learning rate of $10^{-3}$, and a weight decay of 0.05. 
The learning rate follows a half-cycle cosine decay over the 60 global rounds, with four rounds of linear warm-up and a floor of $10^{-6}$.
For FedAdam, the server moments use $\beta_1=0.9$, $\beta_2=0.99$, and $\tau=10^{-3}$.
The FedAdam server learning rate $\eta$ and the FedProx proximal coefficient $\mu$ are selected as described in Section~\ref{sec:hetero}.
Code, configurations, and scripts to reproduce every table, figure, and reported statistic are publicly available at \url{https://github.com/AIHNlab/Fed-ReMasker}.

\section{Results and Discussion}
\subsection{Homogeneous benchmark}
\label{sec:homogeneous}
\begin{figure}[htbp]
\centering
\includegraphics[width=\columnwidth]{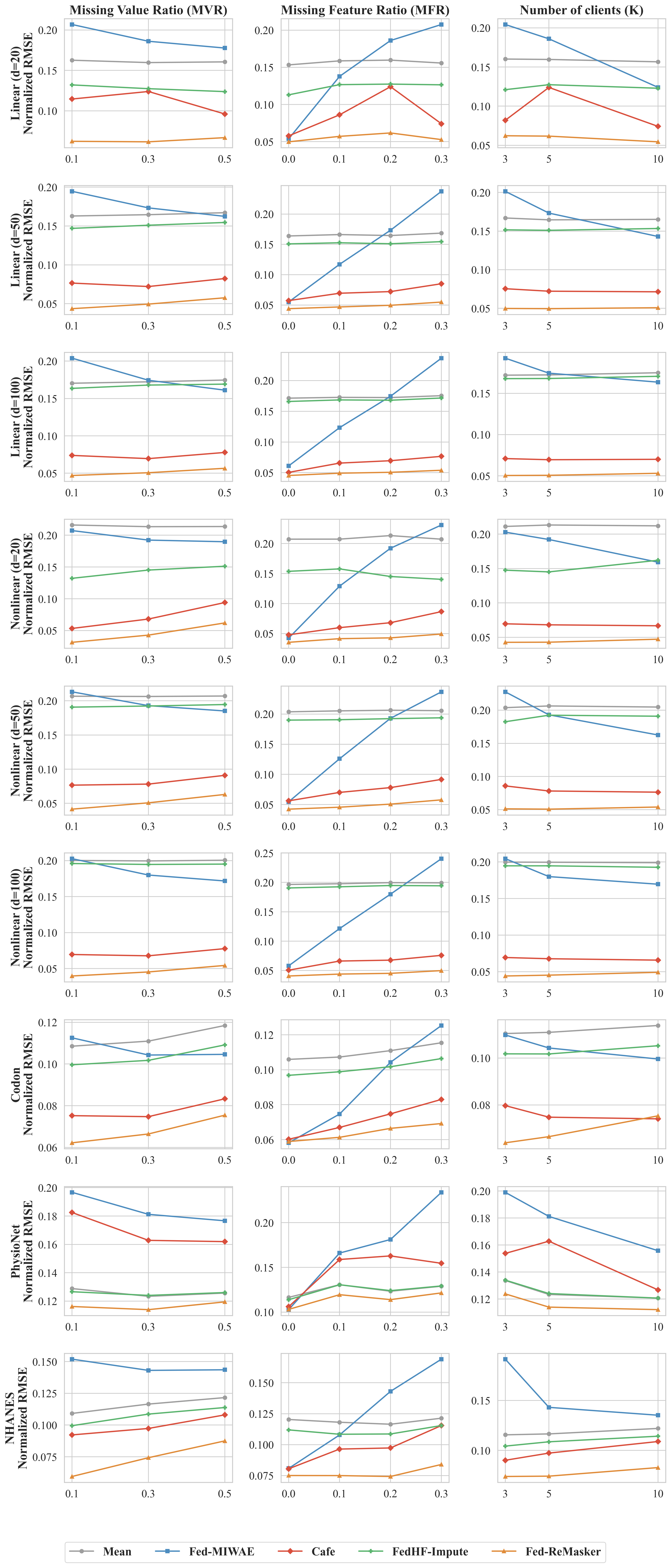}
\caption{Normalized Root Mean Squared Error (NRMSE) over all missing positions of all methods as the Missing Value Ratio (MVR), Missing Feature Ratio (MFR), and number of clients ($K$) vary.
Rows correspond to the six synthetic datasets (linear and nonlinear, at $d \in \{20,50,100\}$) and the three real datasets (Codon, PhysioNet, NHANES), and each column varies one factor while holding the others at $\text{MVR}=0.3$, $\text{MFR}=0.2$, and $K=5$.}
\label{fig1}
\end{figure}

Fig.~\ref{fig1} presents the NRMSE for both synthetic and real datasets as MVR, MFR, and the number of clients vary. 
Fed-ReMasker outperforms the other methods in almost all scenarios, with Cafe being the closest competitor. 
Fed-ReMasker's performance remains relatively stable as MFR and client count increase, whereas Fed-MIWAE degrades sharply as feature-level missingness increases.
Cafe handles low MFR well, but degrades as missing features increase.
PhysioNet has the weakest mean absolute pairwise feature correlations among the three real datasets on the complete pre-masking data ($|r|=0.085$, vs. $0.333$ for Codon and $0.149$ for NHANES).
On PhysioNet, Cafe struggles at high MFR levels and eventually falls behind a basic mean imputation strategy. 
In contrast, Fed-ReMasker maintains its advantage and keeps the lowest NRMSE, even on weakly correlated data.

Table~\ref{tab1} reports the NRMSE averaged across scenarios for each dataset. 
Fed-ReMasker achieves the lowest NRMSE on every dataset for both value-level and feature-level missingness, with average NRMSE values of 0.061 and 0.065, respectively.
Cafe is the second-best method with the corresponding values of 0.079 and 0.090. 
The performance of Fed-ReMasker remains stable as feature dimensionality increases: on the synthetic linear data, its value-level NRMSE decreases by 0.006 as $d$ grows from 20 to 100 (across independently generated datasets with distinct causal structures), and on the nonlinear data it stays between 0.046 and 0.051, while its feature-level NRMSE remains relatively unchanged.
On value-level missingness, Fed-MIWAE (0.118) outperforms FedHF-Impute (0.144) and Mean (0.161). 
However, on feature-level missingness, Fed-MIWAE degrades to the worst result (0.203), falling behind even Mean imputation (0.165) on all nine datasets. 
This contrast highlights that methods that perform well on individual missing values can still struggle to impute entirely missing features, which is the central concern of our work.

\begin{table*}[htbp]
\centering
\caption{Normalized Root Mean Squared Error (NRMSE) results for each method and dataset. The Val and Feat columns report the NRMSE on value-level ($\text{NRMSE}_\text{val}$) and feature-level ($\text{NRMSE}_\text{feat}$) missing positions, respectively.
Values are unweighted arithmetic means of the scenario-level NRMSE values over the 36 $(\text{MFR}, \text{MVR}, K)$ settings per dataset at a single partition seed.
For $\text{NRMSE}_\text{feat}$, 27 settings are included because $\text{MFR}=0$ contains no feature-level missing positions.
The Average row is an unweighted mean across datasets.
\textbf{Bold}: best per dataset; \underline{underlined}: second best.}
\label{tab1}
\setlength{\tabcolsep}{5pt}
\begin{tabular}{lcccccccccc}
\toprule
\textbf{Dataset} & \multicolumn{2}{c}{Mean} & \multicolumn{2}{c}{Fed-MIWAE} & \multicolumn{2}{c}{Cafe} & \multicolumn{2}{c}{FedHF-Impute} & \multicolumn{2}{c}{Fed-ReMasker} \\
\cmidrule(lr){2-3} \cmidrule(lr){4-5} \cmidrule(lr){6-7} \cmidrule(lr){8-9} \cmidrule(lr){10-11}
 & Val & Feat & Val & Feat & Val & Feat & Val & Feat & Val & Feat \\
\midrule
Linear ($d$=20) & 0.156 & 0.158 & 0.111 & 0.204 & \underline{0.074} & \underline{0.074} & 0.117 & 0.132 & \textbf{0.056} & \textbf{0.052} \\
Linear ($d$=50) & 0.165 & 0.166 & 0.117 & 0.208 & \underline{0.071} & \underline{0.077} & 0.151 & 0.154 & \textbf{0.050} & \textbf{0.050} \\
Linear ($d$=100) & 0.173 & 0.173 & 0.119 & 0.218 & \underline{0.064} & \underline{0.076} & 0.168 & 0.169 & \textbf{0.050} & \textbf{0.053} \\
NN ($d$=20) & 0.208 & 0.216 & 0.120 & 0.232 & \underline{0.069} & \underline{0.073} & 0.148 & 0.148 & \textbf{0.046} & \textbf{0.048} \\
NN ($d$=50) & 0.205 & 0.204 & 0.127 & 0.228 & \underline{0.075} & \underline{0.084} & 0.191 & 0.189 & \textbf{0.051} & \textbf{0.053} \\
NN ($d$=100) & 0.198 & 0.199 & 0.122 & 0.226 & \underline{0.065} & \underline{0.076} & 0.192 & 0.194 & \textbf{0.046} & \textbf{0.049} \\
Codon & 0.110 & 0.116 & 0.081 & 0.122 & \underline{0.071} & \underline{0.084} & 0.101 & 0.106 & \textbf{0.065} & \textbf{0.073} \\
PhysioNet & 0.115 & \underline{0.136} & 0.142 & 0.225 & 0.122 & 0.165 & \underline{0.114} & 0.136 & \textbf{0.105} & \textbf{0.128} \\
NHANES & 0.123 & 0.115 & 0.122 & 0.166 & \underline{0.102} & \underline{0.103} & 0.114 & 0.107 & \textbf{0.083} & \textbf{0.075} \\
\midrule
\textbf{Average} & 0.161 & 0.165 & 0.118 & 0.203 & \underline{0.079} & \underline{0.090} & 0.144 & 0.148 & \textbf{0.061} & \textbf{0.065} \\
\bottomrule
\end{tabular}
\end{table*}

Fed-ReMasker reduces value-level NRMSE relative to Cafe, the closest baseline, by $0.018$ (95\% $t$-interval $[0.014, 0.022]$ across the nine datasets) and feature-level NRMSE by $0.026$ (95\% $t$-interval $[0.020, 0.031]$).
Against all baselines jointly, it is the single lowest-error method in $93.2\%$ of the 324 value-level and $96.7\%$ of the 243 feature-level scenarios, not the lowest in 22 and 8 of them, respectively.
These 30 exceptions are small and concentrated: the deficit is at most $0.0058$ NRMSE on value-level and $0.0044$ on feature-level positions, and 21 of the 30 occur at $K=10$.
Fed-MIWAE accounts for 15 of the 22 value-level exceptions, all at $\text{MFR}=0$; once any feature is entirely absent, Fed-ReMasker outperforms it in every one of the 243 remaining scenarios.
When all missing positions are pooled, the NRMSE margins over Fed-MIWAE and FedHF-Impute are $0.083$ and $0.084$ with largely overlapping intervals.
When the positions are separated, the margin over Fed-MIWAE is $0.057$ on value-level and $0.139$ on feature-level positions, whereas over FedHF-Impute it is $0.083$ and $0.084$.

Fed-ReMasker's robustness may be explained by several properties of its design.
First, its Transformer backbone \cite{vaswani2017attention} uses self-attention to capture complex inter-feature relationships, which is particularly beneficial in higher-dimensional feature spaces, where richer feature interactions can be exploited.
Second, because training already relies on masking observed features, imputing entirely unobserved features is a natural extension of the learning objective.
Third, its re-masking objective, shown in the original ReMasker to encourage higher-level representations that remain effective under high missingness \cite{du2024remasker}, carries over to the federated setting.
These properties, already present in the centralized ReMasker, are leveraged in the federated setting, where cross-client aggregation allows the global model to learn feature representations from clients that observe a given feature and transfer this information to clients where it is absent.
This robustness is especially important in federated clinical settings, where centers may use different subsets of features but still need to work within a shared feature space.

\subsection{Robustness to client heterogeneity}
 
\begin{figure*}[htbp]
\centering
\includegraphics[width=1.0\textwidth]{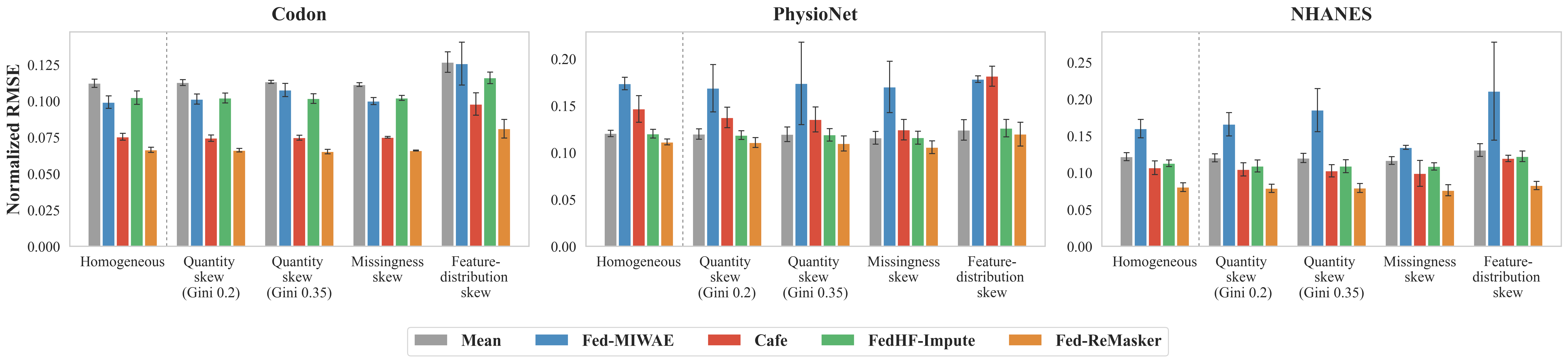}
\caption{
Normalized Root Mean Squared Error (NRMSE) over all missing positions under client heterogeneity. 
Each panel corresponds to one dataset.
A dashed vertical line separates the Homogeneous reference from four heterogeneity conditions, with $\text{MVR}=0.3$, $\text{MFR}=0.2$, and $K=5$ throughout. 
Error bars show $\pm 1$ standard deviation across three client-partition seeds, generated the same way in all conditions.
}
\label{fig2}
\end{figure*}

\begin{table*}[htbp]
\centering
\caption{
Normalized Root Mean Squared Error (NRMSE) for Fed-ReMasker using FedAvg, FedProx, and FedAdam federated optimization strategies on each real dataset under the Homogeneous condition and four heterogeneity conditions. 
Val and Feat report NRMSE on value- and feature-level missing positions. 
Cells are mean $\pm$ standard deviation over three seeds. \textbf{Bold}: best per row; \underline{underlined}: second best, applied within each position type; ties at the reported precision receive the same mark.
}
\label{tab2}
\setlength{\tabcolsep}{3pt}
{\footnotesize
\begin{tabular}{llcccccc}
\toprule
\textbf{Dataset} & \textbf{Scenario} & \multicolumn{2}{c}{\textbf{FedAvg}} & \multicolumn{2}{c}{\textbf{FedProx}} & \multicolumn{2}{c}{\textbf{FedAdam}} \\
\cmidrule(lr){3-4} \cmidrule(lr){5-6} \cmidrule(lr){7-8}
 &  & Val & Feat & Val & Feat & Val & Feat \\
\midrule
Codon & Homogeneous & \textbf{0.065 $\pm$ 0.001} & \textbf{0.068 $\pm$ 0.003} & 0.072 $\pm$ 0.001 & 0.076 $\pm$ 0.003 & \underline{0.067 $\pm$ 0.002} & \underline{0.071 $\pm$ 0.003} \\
Codon & Quantity skew (Gini 0.2) & \textbf{0.065 $\pm$ 0.001} & \textbf{0.067 $\pm$ 0.002} & 0.073 $\pm$ 0.001 & 0.075 $\pm$ 0.002 & \underline{0.067 $\pm$ 0.001} & \underline{0.069 $\pm$ 0.002} \\
Codon & Quantity skew (Gini 0.35) & \textbf{0.064 $\pm$ 0.002} & \textbf{0.066 $\pm$ 0.001} & 0.073 $\pm$ 0.002 & 0.074 $\pm$ 0.002 & \underline{0.066 $\pm$ 0.002} & \underline{0.067 $\pm$ 0.001} \\
Codon & Missingness skew & \textbf{0.065 $\pm$ 0.001} & \textbf{0.067 $\pm$ 0.002} & 0.072 $\pm$ 0.001 & 0.074 $\pm$ 0.003 & \underline{0.068 $\pm$ 0.002} & \underline{0.070 $\pm$ 0.003} \\
Codon & Feature-distribution skew & \textbf{0.069 $\pm$ 0.002} & \textbf{0.093 $\pm$ 0.012} & 0.076 $\pm$ 0.002 & 0.102 $\pm$ 0.016 & \textbf{0.069 $\pm$ 0.002} & \underline{0.097 $\pm$ 0.018} \\
\midrule
PhysioNet & Homogeneous & \textbf{0.106 $\pm$ 0.005} & \textbf{0.116 $\pm$ 0.012} & 0.108 $\pm$ 0.005 & 0.117 $\pm$ 0.012 & \textbf{0.106 $\pm$ 0.005} & \textbf{0.116 $\pm$ 0.013} \\
PhysioNet & Quantity skew (Gini 0.2) & \textbf{0.107 $\pm$ 0.005} & \textbf{0.114 $\pm$ 0.015} & 0.109 $\pm$ 0.006 & 0.116 $\pm$ 0.015 & \underline{0.108 $\pm$ 0.005} & \underline{0.115 $\pm$ 0.014} \\
PhysioNet & Quantity skew (Gini 0.35) & \textbf{0.105 $\pm$ 0.008} & \textbf{0.113 $\pm$ 0.023} & 0.108 $\pm$ 0.008 & \underline{0.115 $\pm$ 0.022} & \underline{0.107 $\pm$ 0.007} & \underline{0.115 $\pm$ 0.025} \\
PhysioNet & Missingness skew & \textbf{0.113 $\pm$ 0.006} & \textbf{0.096 $\pm$ 0.022} & 0.115 $\pm$ 0.006 & \underline{0.097 $\pm$ 0.022} & \textbf{0.113 $\pm$ 0.005} & \underline{0.097 $\pm$ 0.021} \\
PhysioNet & Feature-distribution skew & \underline{0.108 $\pm$ 0.002} & \textbf{0.130 $\pm$ 0.028} & 0.111 $\pm$ 0.003 & \textbf{0.130 $\pm$ 0.028} & \textbf{0.107 $\pm$ 0.004} & 0.133 $\pm$ 0.027 \\
\midrule
NHANES & Homogeneous & \textbf{0.083 $\pm$ 0.004} & \textbf{0.076 $\pm$ 0.018} & 0.090 $\pm$ 0.004 & 0.084 $\pm$ 0.017 & \underline{0.084 $\pm$ 0.004} & \underline{0.077 $\pm$ 0.018} \\
NHANES & Quantity skew (Gini 0.2) & \textbf{0.083 $\pm$ 0.005} & \textbf{0.073 $\pm$ 0.017} & 0.089 $\pm$ 0.005 & 0.079 $\pm$ 0.016 & \underline{0.084 $\pm$ 0.005} & \textbf{0.073 $\pm$ 0.017} \\
NHANES & Quantity skew (Gini 0.35) & \textbf{0.084 $\pm$ 0.004} & \textbf{0.073 $\pm$ 0.017} & 0.089 $\pm$ 0.004 & 0.079 $\pm$ 0.018 & \underline{0.085 $\pm$ 0.004} & \textbf{0.073 $\pm$ 0.017} \\
NHANES & Missingness skew & \textbf{0.085 $\pm$ 0.004} & \textbf{0.062 $\pm$ 0.023} & 0.091 $\pm$ 0.005 & 0.069 $\pm$ 0.022 & \underline{0.086 $\pm$ 0.004} & \underline{0.063 $\pm$ 0.024} \\
NHANES & Feature-distribution skew & \textbf{0.083 $\pm$ 0.004} & \textbf{0.081 $\pm$ 0.017} & 0.090 $\pm$ 0.004 & 0.091 $\pm$ 0.015 & \underline{0.084 $\pm$ 0.004} & \textbf{0.081 $\pm$ 0.016} \\
\bottomrule
\end{tabular}
}
\end{table*}

We next evaluate whether Fed-ReMasker's advantage is maintained under client heterogeneity. 
Fig.~\ref{fig2} compares all methods across the four heterogeneity conditions on the three real datasets. 
Fed-ReMasker achieves the lowest NRMSE across all evaluated heterogeneity conditions.
The relative ordering of methods is also largely preserved from the homogeneous setting, indicating that its advantage is not an artifact of balanced, randomly partitioned clients. 
The feature-distribution skew scenario generally produces the largest degradation, consistent with a global model being pulled toward divergent client distributions.
Fed-ReMasker outperforms every baseline in all 36 value-level scenarios and in at least 35 of 36 feature-level scenarios, beating Cafe, its closest competitor, in all 36.
The same pattern observed on the homogeneous grid is also seen here: the margin over Fed-MIWAE is almost twice as large for feature-level compared to value-level positions ($0.086$ versus $0.044$), whereas the margin over FedHF-Impute remains similar ($0.027$ versus $0.025$).
Equal-client macro NRMSE produces the same method ranking across all heterogeneity scenarios.
Under the strongest quantity skew (Gini $=0.35$, where client sizes differ by $6.8\times$), the largest NRMSE shift between the two weighting schemes across all methods is $0.0062$, whereas Fed-ReMasker changes by only $0.0003$.

Table~\ref{tab2} compares Fed-ReMasker's optimization strategies across FedAvg, FedProx, and FedAdam for value- and feature-level positions.
For FedProx and FedAdam, $\mu$ and $\eta$, respectively, are selected using leave-one-dataset-out hyperparameter selection, which returned $\mu=0.001$ and $\eta=0.01$ for each held-out dataset.
FedAvg achieves the best or tied-best NRMSE in 29 of the 30 cells, with the exception being PhysioNet under feature-distribution skew on value-level positions.
Quantity skew produces only small changes in NRMSE at either Gini level for all three strategies.
Missingness skew leaves the value-level error unchanged on Codon and slightly increases it on PhysioNet and NHANES ($+0.007$ and $+0.002$).
On the same two datasets, it reduces feature-level error by $0.020$ and $0.014$, respectively, plausibly because clients that observe more features provide a stronger imputation signal for the features that others lack.
Feature-distribution skew shows the opposite pattern: it degrades value-level error by at most $0.004$, but increases feature-level error by $0.025$ on Codon, $0.014$ on PhysioNet, and $0.005$ on NHANES.
Overall, the same patterns are observed with FedProx and FedAdam.

\begin{figure}[tp]
\centering
\includegraphics[width=0.48\textwidth]{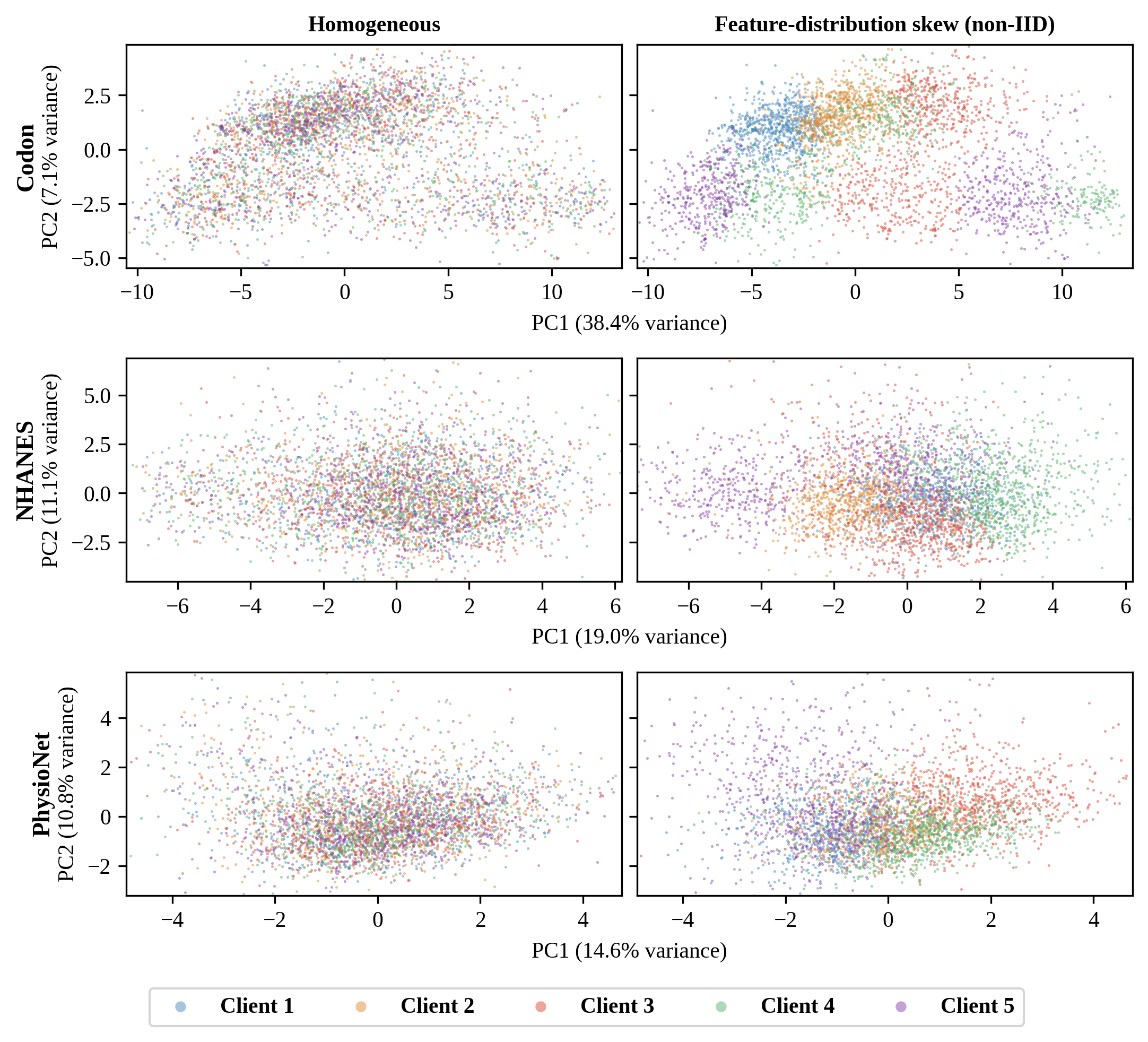}
\caption{Client partitions projected onto the first two principal components of each dataset, computed on the complete pre-masking data with $K=5$.}
\label{fig:pca}
\end{figure}

This difference in sensitivity between value- and feature-level imputation is consistent with the underlying mechanism: value-level positions can be reconstructed from correlations each client observes locally, whereas feature-level positions rely on relationships that can only be learned from clients that observe the feature.
Fig.~\ref{fig:pca} illustrates that the two partitions differ as intended: under the homogeneous split, the clients overlap almost completely, whereas under feature-distribution skew (non-IID), they are visibly separated.
The degree of separation should not be compared directly across datasets because the first two components explain different proportions of the total variance: $45.5\%$, $30.1\%$, and $25.4\%$ for Codon, NHANES, and PhysioNet, respectively.

Feature-level error is also generally less stable across partition seeds, reaching a standard deviation of $\pm0.028$ on PhysioNet under feature-distribution skew scenarios.
Averaged across all conditions and both missingness types, FedProx has $11.3\%$ and $8.7\%$ higher NRMSE than FedAvg on Codon and NHANES but only $1.6\%$  higher NRMSE on PhysioNet.
FedAdam is consistently closer to FedAvg, with relative differences of $3.2\%$, $0.9\%$, $0.8\%$ on Codon, NHANES, PhysioNet, respectively.
On PhysioNet, the largest difference between any two strategies in any cell is $0.004$, within the feature-level seed standard deviation, so the difference is small relative to the observed seed variability.
This is also the dataset on which all methods are closest to mean imputation in Table~\ref{tab1}.
Hyperparameter selection returned the smallest proximal strength in our tested range for FedProx ($\mu=0.001$), suggesting that stronger control of client drift was not beneficial in the settings we tested.
One possible explanation is the shared imputation objective.
All clients solve the same masked-reconstruction task over a shared feature schema, which may already align their updates.
Proximal regularization could therefore constrain local optimization without providing a corresponding benefit from reduced client drift. 
Likewise, FedAdam's server-side adaptivity provides little benefit, consistent with the same explanation: it can rescale the aggregated update but cannot directly address client drift that has already occurred during local training.
Sample-size-weighted averaging is therefore sufficient for Fed-ReMasker under the evaluated conditions, and we retain FedAvg as the default aggregation strategy.

\subsection{Centralized vs. Federated ReMasker}
\label{sec:central}
Table~\ref{tab:central} reports the comparison with the centralized reference of Section~\ref{sec:central_setup}.
Blocks (a) and (b) are averaged over all nine datasets, and block (c) covers the three real datasets over three seeds.
Blocks (a) and (b) overlap at $\text{MFR}=0.2$, $K=5$, but (a) averages over $\text{MVR}\in\{0.1,0.3,0.5\}$ while (b) fixes $\text{MVR}=0.3$.
Averaged over the 126 configurations, Fed-ReMasker's error is $3.0\%$ above the centralized model.

Fed-ReMasker remains within 2.5--3.1\% of centralized ReMasker in relative NRMSE across the MFR sweep, with the largest gap at $\text{MFR}=0$, indicating that feature-level missingness does not substantially increase the cost of federation.
The gap instead grows with client count, from $1.2\%$ at $K=3$ to $3.2\%$ at $K=5$, and $7.2\%$ at $K=10$.
The centralized arm spans $0.002$ NRMSE across the three settings, broadly consistent with the $K$-invariance of the pooled missingness rate. 
Between $K=3$ and $K=5$, the widening reflects a larger improvement in the centralized arm than in the federated arm, whereas from $K=5$ to $K=10$, the centralized error is unchanged and federated error rises from $0.0617$ to $0.0643$. 
Only the latter reflects degradation on the federated side, consistent with each client holding less data.
As the client-count block rests on a single $(\text{MFR}, \text{MVR})$ point and one partition seed, we read this trend as indicative rather than conclusive.
Under the feature-distribution skew condition, the error increases for both centralized and federated models relative to the homogeneous reference, indicating that part of the difficulty of this condition is intrinsic to the resulting missingness pattern rather than a consequence of federation. 
Across the 18 (dataset, condition, seed) scenarios, the mean paired difference between federated and centralized NRMSE is $0.0014$ (95\% $t$-interval $[-0.0077, +0.0105]$ over the three datasets), so we cannot detect an overall advantage from centralizing.
Codon, which has the highest inter-feature correlation, is the only dataset with a consistent advantage (all six scenarios positive, $5.0$--$9.7\%$); NHANES and PhysioNet average $-0.0008$ and $-0.0006$.

\begin{table}[tp]
\centering
\caption{
Fed-ReMasker versus centralized ReMasker trained on the pooled client data, using the same underlying client-level missingness masks and data passes.
$\Delta$\% is the federated minus the centralized NRMSE, relative to the federated NRMSE, so positive values favor centralization.
Values are NRMSE over all missing positions.
Rows average over 27 cells in (a) (9 datasets $\times$ 3 MVR), 9 in (b) (one per dataset), and 3 seeds in (c).
}
\label{tab:central}
\begin{tabular}{lccr}
\toprule
 & Federated & Centralized & $\Delta$\% \\
\midrule
\multicolumn{4}{l}{\textit{(a) Feature-missingness rate ($K=5$)}} \\
\quad MFR $=0$ & 0.0563 & 0.0546 & $+3.1$ \\
\quad MFR $=0.1$ & 0.0617 & 0.0601 & $+2.5$ \\
\quad MFR $=0.2$ & 0.0630 & 0.0612 & $+2.8$ \\
\quad MFR $=0.3$ & 0.0672 & 0.0653 & $+2.8$ \\
\addlinespace
\multicolumn{4}{l}{\textit{(b) Client count (MFR $=0.2$, MVR $=0.3$)}} \\
\quad $K=3$ & 0.0624 & 0.0617 & $+1.2$ \\
\quad $K=5$ & 0.0617 & 0.0597 & $+3.2$ \\
\quad $K=10$ & 0.0643 & 0.0597 & $+7.2$ \\
\addlinespace
\multicolumn{4}{l}{\textit{(c) Heterogeneity (real datasets, $K=5$)}} \\
\quad Codon, Homogeneous & 0.0664 & 0.0603 & $+9.1$ \\
\quad Codon, Feature-distribution skew & 0.0809 & 0.0757 & $+6.5$ \\
\quad NHANES, Homogeneous & 0.0803 & 0.0800 & $+0.3$ \\
\quad NHANES, Feature-distribution skew & 0.0826 & 0.0844 & $-2.2$ \\
\quad PhysioNet, Homogeneous & 0.1112 & 0.1123 & $-1.0$ \\
\quad PhysioNet, Feature-distribution skew & 0.1196 & 0.1197 & $-0.1$ \\
\bottomrule
\end{tabular}
\end{table}

\subsection{Limitations and Future Work}
Our evaluation focuses on MCAR missingness at both the value and feature level, except in the feature-distribution skew condition, where the data-dependent client partition makes feature-level missingness depend on the feature values.
That condition should therefore be read as a combined test of distributional heterogeneity and data-dependent client assignment rather than as a pure distributional-heterogeneity test.
We did not simulate MAR or MNAR value-level mechanisms, nor systematic site-specific feature availability, in which particular centers consistently lack particular variables because of their data-collection protocols.
All evaluated datasets contain at least 9{,}200 samples, leaving performance on the substantially smaller cohorts common in multi-center clinical studies unassessed, and we evaluate quantity, missingness, and feature-distribution skew separately, so their combined effect remains unknown.
Although Fed-ReMasker can impute categorical features, we restricted our evaluation to numerical features for consistency with the baselines, which do not support categorical imputation.
Beyond keeping raw data decentralized, we do not evaluate formal privacy mechanisms such as secure aggregation or differential privacy, nor system costs such as communication and computation; assessing how these trade off against imputation accuracy under realistic deployment is important future work.
We evaluate imputation accuracy directly and do not assess how the imputed data affect downstream predictive or statistical analyses, another important direction for future work.
Finally, the homogeneous grid uses a single client-partition seed, so the paired intervals in Section~\ref{sec:homogeneous} reflect variability across experimental configurations rather than across repeated partitioning, which we quantify only on the heterogeneity subset.
 
\section{Conclusion}
In this work, we present Fed-ReMasker, a federated adaptation of the ReMasker masked autoencoder for tabular imputation under feature-level missingness, where entire features are absent at some centers, a setting largely absent in prior federated imputation research. 
We evaluate Fed-ReMasker through a systematic benchmark that evaluates value- and feature-level missingness separately, across synthetic and real datasets, client counts, and client heterogeneity.
Missingness is MCAR throughout the benchmark except in the feature-distribution skew condition, where the data-dependent client partition induces dependence between the feature-level missingness pattern and the feature values.
Fed-ReMasker achieves the lowest NRMSE on every dataset for both value- and feature-level missing positions, with the largest gains on entirely missing features.
It remains the strongest method under client heterogeneity, with simple FedAvg aggregation achieving the lowest error across almost all evaluated conditions.
It averages 3.0\% above a centralized ReMasker trained on the pooled data, with the gap appearing to widen as the number of clients increases.
By imputing features that are entirely missing at individual centers, even when these features are only partially observed at other centers, Fed-ReMasker enables multi-center studies to jointly leverage datasets with differing feature sets, supporting collaborative analysis across centers without centralizing raw data.

\section*{Acknowledgment}

The BETTER4U project has received funding from the European Union's Horizon Europe Research and Innovation programme under Grant Agreement n° 101080117, by UK Research and Innovation (UKRI) under the UK government's Horizon Europe funding guarantee (grant number 10093560 for QMUL and 10106435 for BiB) and from the Swiss State Secretariat for Education, Research and Innovation (SERI). Views and opinions expressed are, however, those of the author(s) only and do not necessarily reflect those of the European Union.

During the preparation of this work, the authors used Claude (Anthropic) for grammar checking, improving readability, and assisting with code development. 
After using the tool, the authors formally reviewed the content for accuracy and edited it as necessary. 
The authors take full responsibility for all the content of this publication.

\bibliographystyle{IEEEtran}
\bibliography{mybibliography}

\input{consortium_full}

\end{document}

%% file: consortium_full.tex
\section*{BETTER4U Consortium Members}
\footnotesize
\begin{itemize}[leftmargin=1em, itemsep=0pt, parsep=0pt, topsep=2pt, label={}]
\item \textbf{George V. Dedoussis}, Department of Nutrition and Dietetics, School of Health Science and Education, Harokopio University of Athens, 17676 Athens, Greece; Genome Analysis, 17676 Athens, Greece
\item \textbf{Yannis Manios}, Department of Nutrition and Dietetics, School of Health Science and Education, Harokopio University of Athens, 17676 Athens, Greece; European Centre for Obesity, Harokopio University of Athens, 17676 Athens, Greece; Institute of Agri-Food and Life Sciences, Hellenic Mediterranean University Research Centre, 71410 Heraklion, Greece
\item \textbf{Christos Diou}, Department of Informatics and Telematics, School of Digital Technology, Harokopio University of Athens, 17778 Athens, Greece
\item \textbf{Panagiotis Moulos}, Department of Nutrition and Dietetics, School of Health Science and Education, Harokopio University of Athens, 17676 Athens, Greece
\item \textbf{Ioanna Panagiota Kalafati}, Department of Nutrition and Dietetics, School of Health Science and Education, Harokopio University of Athens, 17676 Athens, Greece
\item \textbf{Maria Kafyra}, Department of Nutrition and Dietetics, School of Health Science and Education, Harokopio University of Athens, 17676 Athens, Greece
\item \textbf{Panagiotis Symianakis}, Department of Nutrition and Dietetics, School of Health Science and Education, Harokopio University of Athens, 17676 Athens, Greece
\item \textbf{Eva Karaglani}, Department of Nutrition and Dietetics, School of Health Science and Education, Harokopio University of Athens, 17676 Athens, Greece; European Centre for Obesity, Harokopio University of Athens, 17676 Athens, Greece
\item \textbf{Vasiliki Vavouraki}, Department of Nutrition and Dietetics, School of Health Science and Education, Harokopio University of Athens, 17676 Athens, Greece; Genome Analysis, 17676 Athens, Greece
\item \textbf{Christina Patmiou}, Department of Nutrition and Dietetics, School of Health Science and Education, Harokopio University of Athens, 17676 Athens, Greece
\item \textbf{Paris Kantaras}, Department of Nutrition and Dietetics, School of Health Science and Education, Harokopio University of Athens, 17676 Athens, Greece
\item \textbf{Panagiotis Alimisis}, Department of Informatics and Telematics, School of Digital Technology, Harokopio University of Athens, 17778, Greece
\item \textbf{Anastasios Papamanolis}, Department of Informatics and Telematics, School of Digital Technology, Harokopio University of Athens, 17778, Greece
\item \textbf{Ana Rito}, Centro de Estudos e Investigação em Dinâmicas Sociais e Saúde, 1649-016 Lisbon, Portugal
\item \textbf{Marta Gaspar}, Centro de Estudos e Investigação em Dinâmicas Sociais e Saúde, 1649-016 Lisbon, Portugal
\item \textbf{Matilde Vicente}, Centro de Estudos e Investigação em Dinâmicas Sociais e Saúde, 1649-016 Lisbon, Portugal
\item \textbf{Raquel Henriques}, Centro de Estudos e Investigação em Dinâmicas Sociais e Saúde, 1649-016 Lisbon, Portugal
\item \textbf{Fátima Martins}, Centro de Estudos e Investigação em Dinâmicas Sociais e Saúde, 1649-016 Lisbon, Portugal
\item \textbf{Julie-Anne Nazare}, Centre de Recherche en Nutrition Humaine Rhône-Alpes, Univ-Lyon, CarMeN Laboratory, Inserm U1060, INRAE UMR1397, Université Claude Bernard Lyon 1, Pierre Bénite, France
\item \textbf{Louise Seconda}, Centre de Recherche en Nutrition Humaine Rhône-Alpes, Univ-Lyon, CarMeN Laboratory, Inserm U1060, INRAE UMR1397, Université Claude Bernard Lyon 1, Pierre Bénite, France
\item \textbf{Anestis Dougkas}, Lyfe Institute, Ecully, France and Centre de Recherche en Nutrition Humaine Rhône-Alpes, Univ-Lyon, CarMeN Laboratory, Inserm U1060, INRAE UMR1397, Université Claude Bernard Lyon 1, Pierre Bénite, France
\item \textbf{Klaus Bønnelykke}, Copenhagen Prospective Studies on Asthma in Childhood, COPSAC, Region Hovedstaden 2870 Gentofte, Denmark
\item \textbf{Rebecca Kofod Vinding}, Copenhagen Prospective Studies on Asthma in Childhood, COPSAC, Region Hovedstaden 2870 Gentofte, Denmark
\item \textbf{David Horner}, Copenhagen Prospective Studies on Asthma in Childhood, COPSAC, Region Hovedstaden, 2870 Gentofte, Denmark
\item \textbf{Stephan Kampshoff}, Science Communication Department, European Food Information Council (EUFIC), 1040 Brussels, Belgium
\item \textbf{Darya Silchenko}, Science Communication Department, European Food Information Council (EUFIC), 1040 Brussels, Belgium
\item \textbf{Maria Hassapidou}, Department of Nutritional Sciences \& Dietetics, School of Health Sciences, International Hellenic University, 57400 Thessaloniki, Greece
\item \textbf{Ioannis Pagkalos}, Department of Nutritional Sciences \& Dietetics, School of Health Sciences, International Hellenic University, 57400 Thessaloniki, Greece
\item \textbf{Elena Patra}, Department of Nutritional Sciences \& Dietetics, School of Health Sciences, International Hellenic University, 57400 Thessaloniki, Greece
\item \textbf{Ioannis Ioakeimidis}, Department of Medicine Huddinge (MedH), Karolinska Institutet, 171 77 Stockholm, Sweden
\item \textbf{Anna Ek}, Department of Clinical Science, Intervention and Technology (CLINTEC), Karolinska Institutet, 171 77, Stockholm, Sweden
\item \textbf{Alkyoni Glympi}, Department of Medicine Huddinge (MedH), Karolinska Institutet, 171 77 Stockholm, Sweden
\item \textbf{Eva Schernhammer}, Department of Epidemiology, Center for Public Health, Medical University of Vienna, 1090 Vienna, Austria
\item \textbf{Magdalena Zebrowska}, Department of Epidemiology, Center for Public Health, Medical University of Vienna, 1090 Vienna, Austria
\item \textbf{Constantinos Deltas}, biobank.cy, Center of Excellence in Biobanking and Biomedical Research, University of Cyprus, 2109 Nicosia, Cyprus
\item \textbf{Stavros Gravas}, biobank.cy, Center of Excellence in Biobanking and Biomedical Research, University of Cyprus, 2109 Nicosia, Cyprus
\item \textbf{Panagiota Veloudi}, biobank.cy, Center of Excellence in Biobanking and Biomedical Research, University of Cyprus, 2109 Nicosia, Cyprus
\item \textbf{Alexis Kyriacou}, biobank.cy, Center of Excellence in Biobanking and Biomedical Research, University of Cyprus, 2109 Nicosia, Cyprus
\item \textbf{Apostolos Malatras}, biobank.cy, Center of Excellence in Biobanking and Biomedical Research, University of Cyprus, 2109 Nicosia, Cyprus
\item \textbf{Jaakko Kaprio}, Institute for Molecular Medicine Finland FIMM, HiLIFE, University of Helsinki, FI-00014 Helsinki, Finland
\item \textbf{Teemu Palviainen}, Institute for Molecular Medicine Finland FIMM, HiLIFE, University of Helsinki, FI-00014 Helsinki, Finland
\item \textbf{Gabin Drouard}, Institute for Molecular Medicine Finland FIMM, HiLIFE, University of Helsinki, FI-00014 Helsinki, Finland
\item \textbf{Karri Silventoinen}, Population Research Unit, Faculty of Social Sciences, University of Helsinki, FIN-00014, Helsinki, Finland
\item \textbf{Alvaro Obeso}, Population Research Unit, Faculty of Social Sciences, University of Helsinki, FIN-00014, Helsinki, Finland
\item \textbf{Maira Bes-Rastrollo}, Department of Preventive Medicine and Public Health, University of Navarra; CIBERobn; Navarra Institute for Health Research (IdiSNA), 31008 Pamplona, Spain
\item \textbf{Carmen Sayon-Orea}, Department of Preventive Medicine and Public Health, University of Navarra; CIBERobn; Navarra Institute for Health Research (IdiSNA), 31008 Pamplona, Spain
\item \textbf{Miguel A. Martinez-Gonzalez}, D Department of Preventive Medicine and Public Health, University of Navarra; CIBERobn; Navarra Institute for Health Research (IdiSNA), 31008 Pamplona, Spain
\item \textbf{Cristina Razquin}, Department of Preventive Medicine and Public Health, University of Navarra; CIBERobn; Navarra Institute for Health Research (IdiSNA), 31008 Pamplona, Spain
\item \textbf{Vanessa Bullon-Vela}, Department of Preventive Medicine and Public Health, University of Navarra, 31008 Pamplona, Spain
\item \textbf{Delfien Gryspeerdt}, Department of Public Health and Primary Care, Interuniversity Centre for Health Economics Research (I-CHER), Ghent University, 9000 Ghent, Belgium
\item \textbf{Nick Verhaeghe}, Department of Public Health and Primary Care, Interuniversity Centre for Health Economics Research (I-CHER), Ghent University, 9000 Ghent, Belgium
\item \textbf{Ruben Willems}, Department of Public Health and Primary Care, Interuniversity Centre for Health Economics Research (I-CHER), Ghent University, 9000 Ghent, Belgium
\item \textbf{Lieven Annemans}, Department of Public Health and Primary Care, Interuniversity Centre for Health Economics Research (I-CHER), Ghent University, 9000 Ghent, Belgium
\item \textbf{Aleksandra Luszczynska}, Institute of Psychology, CARE-BEH Center for Applied Research on Health Behavior and Health, SWPS University, 03815 Warsaw, Poland
\item \textbf{Paulina Krzywicka}, Institute of Psychology, CARE-BEH Center for Applied Research on Health Behavior and Health, SWPS University, 03815 Warsaw, Poland
\item \textbf{Zofia Szczuka}, Institute of Psychology, CARE-BEH Center for Applied Research on Health Behavior and Health, SWPS University, 03815 Warsaw, Poland
\item \textbf{Hanna Zaleskiewicz}, Institute of Psychology, CARE-BEH Center for Applied Research on Health Behavior and Health, SWPS University, 03815 Warsaw, Poland
\item \textbf{Anna Kornafel}, Institute of Psychology, CARE-BEH Center for Applied Research on Health Behavior and Health, SWPS University, 03815 Warsaw, Poland
\item \textbf{Anders Eriksson}, Institute of Genomics, University of Tartu, 51010 Tartu, Estonia
\item \textbf{Elisabeth Thiering}, Institute of Epidemiology, Helmholtz Zentrum München, German Research Center for Environmental Health, 85764, Neuherberg, Germany
\item \textbf{Marie Standl}, Institute of Epidemiology, Institute of Epidemiology, Helmholtz Zentrum München, German Research Center for Environmental Health, 85764, Neuherberg, Germany
\item \textbf{Tamara Schikowski}, IUF---Leibniz Research Institute for Environmental Medicine
\item \textbf{Gunda Herberth}, Department of Environmental Immunology, Helmholtz Centre for Environmental Research -- UFZ
\item \textbf{Dorret I. Boomsma}, Department of Biological Psychology, VU Amsterdam, 1081 BT, Amsterdam, Netherlands
\item \textbf{René Pool}, Department of Biological Psychology, VU Amsterdam, 1081 BT, Amsterdam, Netherlands
\item \textbf{Susanne Bruins}, Department of Biological Psychology, VU Amsterdam, 1081 BT, Amsterdam, Netherlands
\item \textbf{Euan Woodward}, The European Association for the Study of Obesity, D02 N820, Dublin, Ireland
\item \textbf{Lisa Heggie}, The European Association for the Study of Obesity, D02 N820, Dublin, Ireland
\item \textbf{Sheree Bryant}, The European Association for the Study of Obesity, D02 N820, Dublin, Ireland
\item \textbf{Rodessa May Marquez}, Department of Innovation and Digitalisation in Law, University of Vienna, 1010 Vienna, Austria
\item \textbf{Olga Startseva}, Department of Innovation and Digitalisation in Law, University of Vienna, 1010 Vienna, Austria
\item \textbf{Nikolaus Forgó}, Department of Innovation and Digitalisation in Law, University of Vienna, 1010 Vienna, Austria
\item \textbf{Eran Segal}, Department of Computer Science and Appliance and Applied Mathematics, Weizmann Institute of Science, 7610001 Rehovot, Israel
\item \textbf{Adina Weinberger}, Department of Computer Science and Appliance and Applied Mathematics, Weizmann Institute of Science, 7610001 Rehovot, Israel
\item \textbf{Vera Stavroulaki}, Wings ICT Solutions Information \& Communication Technologies S.A., 17121 Athens, Greece
\item \textbf{Vangelis Argoudelis}, Wings ICT Solutions Information \& Communication Technologies S.A., 17121 Athens, Greece
\item \textbf{Panagiotis Demestichas}, Wings ICT Solutions Information \& Communication Technologies S.A., 17121 Athens, Greece
\item \textbf{Danai Malti}, Wings ICT Solutions Information \& Communication Technologies S.A., 17121 Athens, Greece
\item \textbf{Gianna Karanasiou}, Wings ICT Solutions Information \& Communication Technologies S.A., 17121 Athens, Greece
\item \textbf{Dimitris Plakas}, Wings ICT Solutions Information \& Communication Technologies S.A., 17121 Athens, Greece
\item \textbf{Nikos Sintoris}, Wings ICT Solutions Information \& Communication Technologies S.A., 17121 Athens, Greece
\item \textbf{Vassilios Fanos}, Department of Surgical Sciences, University of Cagliari, 09042 Monserrato, Italy
\item \textbf{Angelica Dessì}, Department of Surgical Sciences, University of Cagliari, 09042 Monserrato, Italy
\item \textbf{Luigi Atzori}, Department of Biomedical Sciences, University of Cagliari, 09042 Monserrato, Italy
\item \textbf{Cristina Piras}, Department of Biomedical Sciences, University of Cagliari, 09042 Monserrato, Italy
\item \textbf{Antonio Noto}, Department of Biomedical Sciences, University of Cagliari, 09042 Monserrato, Italy
\item \textbf{Patrizia Baire}, Department of Surgical Sciences, University of Cagliari, 09042 Monserrato, Italy
\item \textbf{Matteo Mauri}, Department of Surgical Sciences, University of Cagliari, 09042 Monserrato, Italy
\item \textbf{Karolina Krystyna Kopeć}, Department of Surgical Sciences, University of Cagliari, 09042 Monserrato, Italy
\item \textbf{Cristinel Gheorghiu}, BIOCLINICA SA, 300358, Timisoara, Romania
\item \textbf{Anastasios Delopoulos}, Department of Electrical and Computer Engineering, Aristotle University of Thessaloniki, 54636 Thessaloniki, Greece
\item \textbf{Ioannis Sarafis}, Department of Electrical and Computer Engineering, Aristotle University of Thessaloniki, 54636 Thessaloniki, Greece
\item \textbf{Alexandros Papadopoulos}, Department of Electrical and Computer Engineering, Aristotle University of Thessaloniki, 54636 Thessaloniki, Greece
\item \textbf{Chrysa Episkopou}, Department of Electrical and Computer Engineering, Aristotle University of Thessaloniki, 54636 Thessaloniki, Greece
\item \textbf{Dimitrios Aletras}, Department of Electrical and Computer Engineering, Aristotle University of Thessaloniki, 54636 Thessaloniki, Greece
\item \textbf{Terho Lehtimäki}, Department of Clinical Chemistry, Tampere University Hospital and Tampere University, 33520 Tampere, Finland
\item \textbf{Nina Mononen}, Department of Clinical Chemistry, Tampere University Hospital and Tampere University, 33520 Tampere, Finland
\item \textbf{Pashupati P Mishra}, Department of Clinical Chemistry, Tampere University Hospital and Tampere University, 33520 Tampere, Finland
\item \textbf{Binisha H Mishra}, Department of Clinical Chemistry, Tampere University Hospital and Tampere University, 33520 Tampere, Finland
\item \textbf{Leo-Pekka Lyytikäinen}, Department of Clinical Chemistry, Tampere University Hospital and Tampere University, 33520 Tampere, Finland
\item \textbf{Stavroula Mougiakakou}, ARTORG Center for Biomedical Engineering Research/AI in Health and Nutrition, University of Bern, 3008 Bern, Switzerland
\item \textbf{Ioannis Papathanail}, ARTORG Center for Biomedical Engineering Research/AI in Health and Nutrition, University of Bern, 3008 Bern, Switzerland
\item \textbf{Rooholla Poursoleymani}, ARTORG Center for Biomedical Engineering Research/AI in Health and Nutrition, University of Bern, 3008 Bern, Switzerland
\item \textbf{Lubnaa Abdur Rahman}, ARTORG Center for Biomedical Engineering Research/AI in Health and Nutrition, University of Bern, 3008 Bern, Switzerland
\item \textbf{Nick Martin}, Department of Genetics and Computational Biology, QIMR Berghofer Medical Research Institute, 4029 Brisbane, Australia
\item \textbf{Scott Gordon}, Department of Genetics and Computational Biology, QIMR Berghofer Medical Research Institute, 4029 Brisbane, Australia
\item \textbf{Gillian Santorelli}, Bradford Institute for Health Research, Bradford Teaching Hospitals NHS Foundation Trust, BD9 6RJ Bradford, United Kingdom
\item \textbf{Ellena Badrick}, Bradford Institute for Health Research, Bradford Teaching Hospitals NHS Foundation Trust, BD9 6RJ Bradford, United Kingdom
\item \textbf{John Wright}, Bradford Institute for Health Research, Bradford Teaching Hospitals NHS Foundation Trust, BD9 6RJ Bradford, United Kingdom
\item \textbf{Amy Hough}, Bradford Institute for Health Research, Bradford Teaching Hospitals NHS Foundation Trust, BD9 6RJ Bradford, United Kingdom
\item \textbf{Eirini Marouli}, William Harvey Research Institute, Queen Mary University of London, EC1M 6BQ London, United Kingdom
\end{itemize}
\normalsize